%% file: main.tex
\documentclass[10pt,twocolumn,letterpaper]{article}

\usepackage[pagenumbers]{wacv} % To force page numbers, e.g. for an arXiv version

\usepackage{graphicx}
\usepackage{amsmath}
\usepackage{amssymb}
\usepackage{booktabs}
\usepackage{multirow}

\usepackage{pifont}
\usepackage{xcolor}
\usepackage{longtable}
\usepackage{subcaption}
\usepackage{dblfloatfix}
\usepackage{float}
\usepackage{booktabs}
\usepackage{bigstrut}
\usepackage{multirow}
\usepackage{color}
\usepackage{caption}
\usepackage{fancyhdr}
\usepackage{listings} %if lstlistings is used
\usepackage{changepage}
\usepackage{pgfplots}
\pgfplotsset{compat=1.8}
\usepackage{balance}
\usepackage{enumitem}
\usepackage{pifont}% http://ctan.org/pkg/pifont
\usepackage[pagebackref,breaklinks,colorlinks]{hyperref}
\usepackage[capitalize]{cleveref}

\crefname{section}{Sec.}{Secs.}
\Crefname{section}{Section}{Sections}
\Crefname{table}{Table}{Tables}
\crefname{table}{Tab.}{Tabs.}

\def\wacvPaperID{21} % *** Enter the WACV Paper ID here
\def\confName{WACV}
\def\confYear{2025}

\begin{document}

\title{Generating Realistic Forehead-Creases for User Verification via Conditioned Piecewise Polynomial Curves}

% \author{Abhishek Tandon\textsuperscript{1}, Geetanjali Sharma\textsuperscript{1}, Gaurav Jaswal \textsuperscript{2}, Aditya Nigam \textsuperscript{1}, Raghavendra Ramachandra\textsuperscript{3}\\
% \textsuperscript{1}IIT Mandi, India, 
% \textsuperscript{2}TIH - IIT Mandi, India, 
% \textsuperscript{3}NTNU, Norway
% }

\author{Abhishek Tandon\textsuperscript{1}, Geetanjali Sharma\textsuperscript{1}, Gaurav Jaswal \textsuperscript{2}, Aditya Nigam \textsuperscript{1}, Raghavendra Ramachandra\textsuperscript{3}\\
\textsuperscript{1}Indian Institute of Technology  Mandi, India \\
\textsuperscript{2}Technology Innovation Hub - Indian Institute of Technology, Mandi\\
\textsuperscript{3}Norwegian University of Science and Technology (NTNU), Norway
}

% \author{Abhishek Tandon\textsuperscript{1}\quad Geetanjali Sharma\textsuperscript{1}\quad Gaurav Jaswal \textsuperscript{2}\\Aditya Nigam \textsuperscript{1}\quad Raghavendra Ramachandra\textsuperscript{3}\\
% \textsuperscript{1}Indian Institute of Technology  Mandi, India \\
% \textsuperscript{2}Technology Innovation Hub - Indian Institute of Technology, Mandi\\
% \textsuperscript{3}Norwegian University of Science and Technology (NTNU), Norway
% }

% \author{Abhishek Tandon\\
% \and
\maketitle

%%%%%%%%% ABSTRACT

\input{sections/0_abstract}
\input{sections/1_introduction}

\input{sections/2_related_work}
\input{sections/4_method}

\input{sections/5_experimental_setup}

\input{sections/6_results}

\input{sections/7_conclusion}

%%%%%%%%% REFERENCES
{\small
\bibliographystyle{ieee_fullname}
\bibliography{egbib}
}

\end{document}

%% file: sections/0_abstract.tex
\begin{abstract}

We propose a trait-specific image generation method that models forehead creases geometrically using B-spline and B\'ezier curves. This approach ensures the realistic generation of both principal creases and non-prominent crease patterns, effectively constructing detailed and authentic forehead-crease images. These geometrically rendered images serve as visual prompts for a diffusion-based Edge-to-Image translation model, which generates corresponding mated samples. The resulting novel synthetic identities are then used to train a forehead-crease verification network.
To enhance intra-subject diversity in the generated samples, we employ two strategies: (a) perturbing the control points of B-splines under defined constraints to maintain label consistency, and (b) applying image-level augmentations to the geometric visual prompts, such as dropout and elastic transformations, specifically tailored to crease patterns.
By integrating the proposed synthetic dataset with real-world data, our method significantly improves the performance of forehead-crease verification systems under a cross-database verification protocol. 

\end{abstract}

%% file: sections/1_introduction.tex
\section{Introduction} \label{label:introduction}
Biometrics is a technology that leverages unique physiological or behavioral traits—such as fingerprints, facial features, or voice patterns—to verify an individual's identity. It is primarily used in security and access control, offering a more secure and reliable alternative to traditional passwords or PINs. With its combination of convenience and robust fraud prevention capabilities, biometrics is increasingly being integrated into smartphones, banking systems, and border security applications, driving its widespread adoption across various sectors.

Biometric traits are widely used for identity verification, relying on unique physiological or behavioral characteristics specific to each individual. Established modalities such as fingerprints, facial recognition, iris patterns, and voice analysis are commonly employed in various applications, including border control, due to their high level of accuracy—particularly when captured under controlled conditions. These technologies have proven to be reliable and effective in ensuring secure authentication. In addition to these well-established methods, researchers are exploring emerging biometric traits to address limitations and expand the range of secure identification options. One such innovation is the use of forehead creases, which analyze the distinctive patterns formed by the creases on an individual’s forehead. This approach offers a valuable alternative in scenarios where traditional biometric modalities may be less effective, such as when fingerprints are damaged, facial features are obstructed, or environmental factors affect voice recognition. By introducing unconventional traits like forehead creases, biometric systems gain greater versatility and resilience, enabling robust authentication even in challenging conditions.

The use of forehead-creases as biometric identifiers was first studied in \cite{bharadwaj2022mobile} and \cite{10593960}. Their methods yielded strong verification performance using a closed-set matching protocol, while failed significantly in open-set matching \cite{bharadwaj2022mobile}. However, \cite{tandon2024synthetic} introduced a more challenging cross-database verification protocol, revealing the need for better generalization. The advent of generative models, such as Diffusion Models \cite{rombach2022high} and GANs \cite{goodfellow2020generative}, offers promising avenues for addressing these challenges. By generating high-quality synthetic biometric data, these models can augment existing datasets, enabling more robust training. When combined with real-world data, this approach has the potential to significantly enhance the generalization capabilities of recognition systems, making them more effective in diverse and real-world scenarios.

\input{figures_tex/main_figure}

The study in \cite{tandon2024synthetic} addresses the challenge of limited forehead-crease image data but falls short as their proposed subject-agnostic approach lacks intra-class diversity. While GANs \cite{goodfellow2020generative} are prone to mode collapse, training diffusion models on biometric modalities with limited data can result in overfitting \cite{moon2022fine} and identity leakage. For crease-based biometric modalities, we assert to use a generative model that learns from the edge and texture domains to accurately capture fine-grained crease patterns.
Recent advancements have explored generating unique synthetic IDs using Bézier curves, as demonstrated in studies on palmprints \cite{zhao2022bezierpalm}, \cite{shen2023rpg}, and \cite{jin2024pce}. These approaches highlight the potential of geometry-driven methods in synthesizing biometric traits, paving the way for their application in forehead-crease generation. A similar strategy could address the specific challenges associated with crease-based biometrics by capturing subtle details while maintaining diversity and identity fidelity.

In this paper, we present a trait-specific biometric image generation technique designed to address the challenges of modalities with limited data availability. The proposed approach leverages B-splines and Bézier curves to ensure diversity, realism, and adaptability across various capture conditions of forehead biometrics. The key contributions of this work are summarized as follows \footnote{ Code and Synthetic Datasets: \href{https://github.com/abhishektandon/bspline-fc}{github.com/abhishektandon/bspline-fc}}:

\begin{itemize} [leftmargin=*,noitemsep, topsep=0pt,parsep=0pt,partopsep=0pt]

    \item We propose an algorithm based on a dynamically generated grid mask that geometrically models forehead-crease patterns using B-splines and Bézier curves, with well-defined constraints, to synthesize unique \textit{visual prompts}. 
    
    \item These \textit{unique visual prompts} are used as input during inference in a diffusion-based edge-to-image translation network, which is trained with dilated self-quotient edge maps and their corresponding forehead-crease image pairs. This process generates novel synthetic forehead biometrics IDs.

    \item To further enhance the intra-subject diversity of synthetic IDs, we perturb the control points of the parametric curves and apply augmentations to the visual prompts, ensuring the generation of diverse mated samples while maintaining label consistency.

\end{itemize}

In subsequent sections, we discuss the related literature of crease-based biometric data generation and recognition in Section \ref{label:related_work}. We elaborate our proposed method in Section \ref{label:methodology}, discuss the experimental protocol in Section \ref{label:experimental_setup}, and present our results with discussions in Section \ref{label:results} for qualitative synthetic data generation and cross-database verification performances.

%% file: figures_tex/main_figure.tex
\definecolor{lightpurple}{rgb}{0.8, 0.7, 1}
\definecolor{lightblue}{rgb}{0.7, 0.85, 1}

\begin{figure*}[!ht]
\centering
\includegraphics[width=1.0\textwidth]{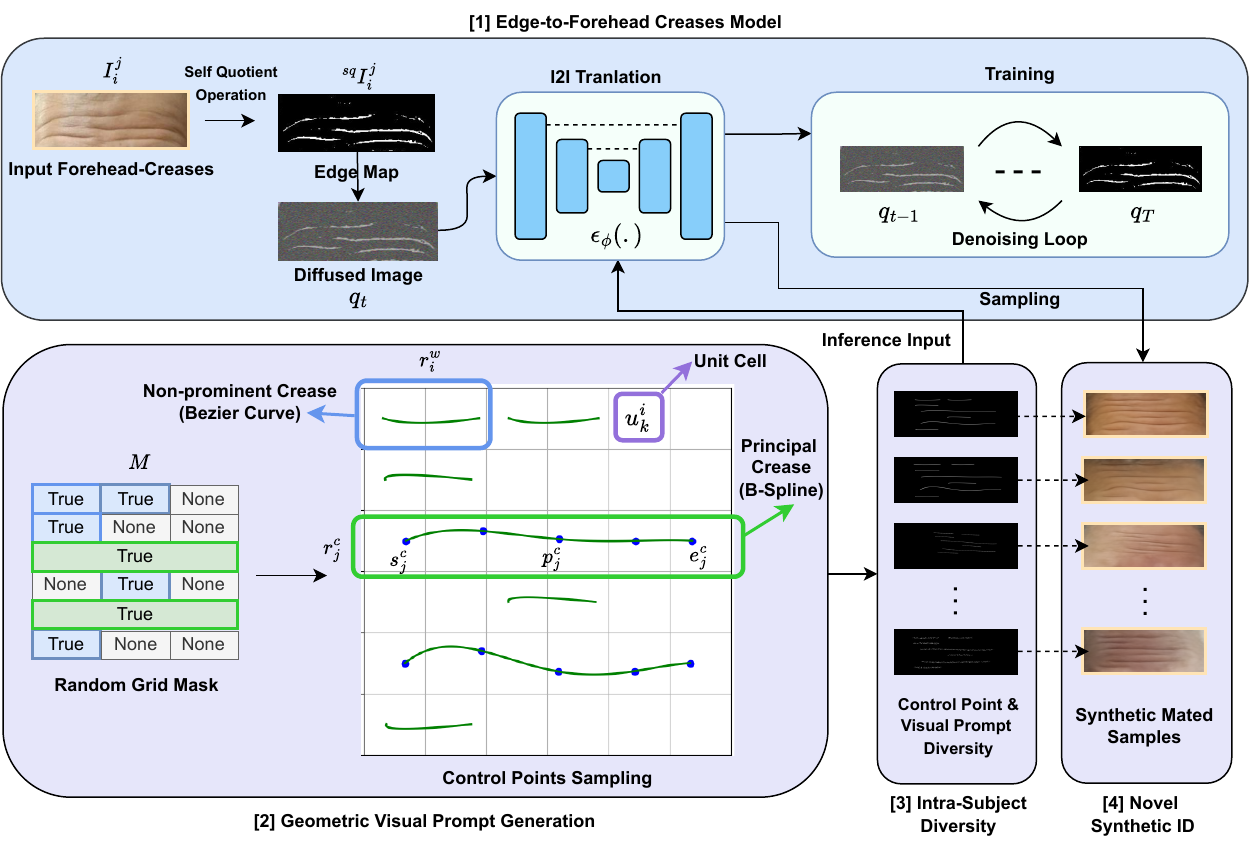}
\caption{Proposed Method: Geometrically rendered \textit{visual prompts} use an Edge-to-Forehead Creases model to generate novel synthetic IDs. Principal creases are rendered using B-spine curves on $j^{th}$ row $r^c_j$ of a $6 \times 6$ grid with start, control and end points: $s^c_j$, $p^c_j$, and, $e^c_j$, while non-prominent creases are plotted using B\'ezier curves on $r^w_j$ formed by merging two consecutive cells $u_k^i$. The spatial position for each crease is decided by a random grid mask $M$. The rounded box with a \textcolor{lightblue}{light-blue} background (top) indicates the training phase, while the remaining three in a \textcolor{lightpurple}{medium-purple} background (bottom) denotes the inference stage pipeline. Best viewed in color.} 
\label{fig_main_figure}
\end{figure*}

%% file: sections/2_related_work.tex
\section{Related Work} \label{label:related_work}
BézierPalm \cite{zhao2022bezierpalm} visualizes intermediate feature maps of palmprint ROI images using a squeeze-and-excite operation, highlighting the importance of crease patterns for palmprint recognition. Motivated by these findings, they generate novel palmprint identities using Bézier curves. To enhance intra-subject diversity, they superimpose the generated creases onto random images from the ImageNet \cite{deng2009imagenet} dataset. RPG-Palm \cite{shen2023rpg} further refines this process by using a threshold \cite{jia2008palmprint} to filter overlapping identities and proposes a framework where mated samples are generated using their Image-to-Image (I2I) GAN model along with an identity loss function. PCE-Palm \cite{jin2024pce} introduces a learnable module based on Gaussian-MFRAT kernels \cite{jia2008palmprint} to enhance texture patterns in the BézierPalm images. Recently, GenPalm \cite{10744493} generate a synthetic palmprint biometrics database with 12,000 unique identities using their proposed ID-ControlNet module conditioned on image quality of palmprint ROIs via text prompts. 

Our work closely follows the literature on crease-based biometric image generation but also draws inspiration from the medical field, where experts propose surgical procedures for the treatment of forehead creases \cite{anido2017tailored} \cite{alhallak2024optimizing}. Due to anatomical variations and differing forehead muscle activity in individuals, physicians recommend dividing the forehead region into a "12-zone map" \cite{anido2017tailored} to accurately define injection points for effective treatment. Building on this concept, we propose a geometric model for forehead creases on a grid-based canvas, using carefully defined parameters to capture all possible variations of forehead creases through B-splines and Bézier curves, subject to well-established constraints.

%% file: sections/4_method.tex
\section{Proposed Methodology} \label{label:methodology}

Figure \ref{fig_main_figure} presents the block diagram of the proposed method for reliable forehead-creases verification. The approach begins by training a diffusion-based Image-to-Image translation model \cite{li2023bbdm} on real forehead-creases images \cite{bharadwaj2022mobile} and their corresponding edge maps, generated through a self-quotient operation followed by dilation. Once the model is trained, it is used to generate novel synthetic IDs by inferring unique visual prompts. These visual prompts are created by geometrically simulating forehead creases using B-spline and Bézier curves on a grid-based canvas. The spatial positioning of the creases is defined by a randomly generated grid mask and the control points of the curves. To enhance diversity within the synthetic mated samples, the visual prompts are further augmented, and the control parameters of the curves are perturbed while maintaining the identity consistency of the generated samples.

\subsection{Edge Extraction} Given an $i^{th}$ ROI forehead-creases image $I^{j}_{i}$ of a subject $j$ from a biometric database \cite{bharadwaj2022mobile} consisting of $d$ enrolled IDs, we extract the crease patterns using image processing. Each image is smoothened using Gaussian blur (kernel size of $15 \times 15$, with standard deviation $\sigma$ = $30.0$) and enhanced using a self-quotient operation which is further binarized (Otsu’s thresholding), dilated ($3 \times 3$ kernel, $1$ iteration) and inverted to obtain an edge map denoted as ${ }^{sq} I_i^j$ (Figure \ref{fig_main_figure} and \ref{fig:edge_extraction}). 

\input{figures_tex/edge_extraction}

The self-quotient operation enhances crease patterns and is less sensitive to external factors such as irregular lighting \cite{kumar2015recovering}, however, excessive dilation can lead to the loss of fine crease details. Each dilated-self quotient image ${ }^{sq} I_i^j$ and its corresponding forehead-creases image $I^{j}_{i}$ act as training image pairs for a diffusion-based Image-to-Image translation model \cite{li2023bbdm}. The ROI segmentation strategy to obtain $I^{j}_{i}$ from masked-face images is the same as described in \cite{bharadwaj2022mobile}. \label{sec:self_q}

\subsection{Edge-to-Forehead Creases Generative Network} We utilize a diffusion-based Image-to-Image Translation network BBDM \cite{li2023bbdm} that takes $({ }^{sq} I_i^j, \, I^{j}_{i})$ as training image pairs. It defines the diffusion trajectory directly between the edge map and target image pairs. The overall training objective is defined as:
\begin{equation}
    \mathbb{E}_{{q}_0, {gt}, {\epsilon}}[|| \epsilon_{\phi}(q_{t}, t) - (\sqrt{\delta_{t}}\epsilon + r_t(y - x_{0})) ||^2]
\end{equation}
where $y$ and $x_0$ are training image pairs for the diffusion model, and $q_t$ is the noised edge map at time $t$ according to the equation:
\begin{equation}
     q_t = (1 - r_t)x_0 + r_t \cdot y + \sqrt\delta_t\epsilon
\end{equation}
The trajectory ratio $r_t$ is $\frac{t}{T}$, variance $\delta_{t} = 2 \cdot (r_{t} - r^{2}_{t})$ and maximum time step $T = 1000$. A UNet-based model \cite{dhariwal2021diffusion} $\epsilon_{\phi}(.)$  predicts the noise added to the edge map at time $t$ during training. During inference, it can generate multiple forehead-creases images given unseen edge maps as input. We refer to this pre-trained model as $Edge2FC$ in later sections.
  
\subsection{Geometric Visual Prompts Generation} \label{geometric_vis_prompt_gen} 

B\'ezier curves are parametric curves defined by a start point, an endpoint and one or more control points. However, they only allow global control, i.e. shape of the whole curve changes if one of the control points is altered. B-spline curves, on the other hand, offer localized control by dividing the curve into segments using \textit{knot vectors} \cite{patrikalakis2002shape}. Our goal is to simulate forehead-creases with well-defined constraints geometrically. The resulting visual prompts act as control inputs to the pre-trained $Edge2FC$ model. To achieve this, we first define the nomenclature based on which the semantic or spatial orientation of the creases is defined.

    \subsubsection{Principal Creases:} Based on observations from the real forehead-creases images database \cite{bharadwaj2022mobile}, we identify certain creases are prominent, i.e. horizontally span a major length of the forehead, and can vary from upto 6 in number. We define these creases as \textit{principal creases}.
    
    \subsubsection{Non-prominent Creases:} We refer to creases that are smaller in length and scattered between two principal creases or other similar creases as \textit{non-prominent creases}. B-splines very well capture the non-symmetrical nature of forehead-creases. Principal creases are therefore rendered using b-splines, while non-prominent creases, which are less complex in shape, are plotted using b\'ezier curves as shown in Figure \ref{fig_main_figure}.

\subsubsection{Dynamic Grid Mask Generation} \label{vp_gen}
The drawing canvas is divided into a $6 \times 6$ grid, where each row either contains a principal crease or non-prominent creases, which is decided by a randomly generated grid mask. A principal crease always spans over an entire row, covering each $1 \times 1$ unit cell ${u^i_k}$ of a row $r_i$. For non-prominent creases, a row is split into three $1 \times 2$ dimension cells, merging two adjacent unit cells into one.

We sample the number of rows from a uniform distribution $r \sim U(3,\, 6)$, i.e. a subset rows of the canvas grid. Among the chosen rows, principal creases are sampled uniformly $r^c \sim U(1,\, r)$, and consequently, the number of rows $r^{w} $ with non-prominent creases are sampled from $U(1,\, r - r^c)$. For a given row $r^{w}_i$, each of the three merged cells, may contain $w \sim U(1,\, 3)$ non-prominent creases. This can effectively be represented as a dynamic array $M = \{m_1, m_2, \dots, m_{6}\}$ such that: 

\begin{equation}
    \centering
    m_i \in \left\{
    \begin{array}{ll}
      True; & \quad \text{for a principal crease} \\
      W; & \quad \text{for a non-prominent crease} \\
      None; & \quad \text{for no crease}
    \end{array}
    \right.
\end{equation}

where $W = \{w_1, w_2, w_3 \}$ is a sub-array and $w_i \in \{True,\, None\}$.
If $m_i$ takes the value $True$, control points for a principal crease will be sampled, and the curve will be represented by a B-spline curve of degree $d_c \sim U(3,\, 4)$. The coefficients of the curve is estimated using $splrep$ function of $scipy$ library\footnote{\href{https://docs.scipy.org/doc/scipy/reference/interpolate.html}{https://docs.scipy.org/doc/scipy/reference/interpolate.html}}. If $m_i$ is a sub-array $W$, it is represented by a degree $d_w = 2$ B\'ezier curve\footnote{\href{https://bezier.readthedocs.io/en/stable/}{https://bezier.readthedocs.io/en/stable/}}. $M$ is padded with $None$ values at the start and end if $r < 6$ such that the creases are plotted at the center of the canvas.

\subsubsection{Sampling Control Points}
After a dynamic mask is generated randomly, the parameters of the curves are sampled and a curve is interpolated. Since the b-spline representation of a principal crease $c$ always spans a complete row $r^c_i$, the start and end points ($s^{c}_i$ and $e^{c}_i$) of the curve are sampled from the center of the first and last unit cell (i.e. $u^i_1$ and $u^i_6$). 

The control points denoted as
$P$ = $\{P_1, \, P_2, ..., P_{d_c + 2}\}$
where $P_j := (x_j, \, y_j)$. First, these points are uniformly placed on the baseline axis, i.e. the line joining $s^{c}_i$ and $e^{c}_i$. Then, the x and y-coordinates of the control points are marginally perturbed by $m$ ($ 0.3 \le m \le 0.6$):
\begin{equation}
    P_j = P_j + \Delta^{c}_{j}; \label{eq_perturb1}
\end{equation}
\begin{equation}
    \Delta^{c}_{j} = \text{random}(-m, \, m) \label{eq_perturb2}
\end{equation}

For non-prominent creases, a simple bézier curve with a start, end and a single control point is chosen, and its parameters are sampled in a similar manner, except it spans across a $1 \times 2$ cell, instead of an entire row. The randomly generated dynamic mask and the control points sampling technique geometrically models the patterns in real forehead-creases, resulting into unique synthetic IDs.

\subsection{Intra-Subject Diversity}
We observe a lack of intra-subject diversity in the mated samples generated from the Edge2FC model (evident from \textit{"BSpline-FC"} dataset in Table \ref{table:fid}). To solve this, we carry out two techniques:
\subsubsection{Control Point Diversity (CPD):} As described in Section \ref{geometric_vis_prompt_gen}, we use the localized control property of b-splines to achieve intra-subject diversity with label consistency. For a given synthetic ID and its parameter array corresponding to its dynamic grid mask, we perturb the array by adding a small random noise to all its parameters (equation \ref{eq_perturb1} and \ref{eq_perturb2}) and generate the mated samples. Note that this does not alter the ID as the location of the creases is already fixed in the grid mask for a given ID.
\subsubsection{Visual Prompt Diversity (VPD):} We identify suitable augmentations that directly mould the visual prompt in the input space so that gets reflected in the generated output. These augmentations range from dropout, elastic transform, piecewise affine, perspective transforms, etc., based on the $imgaug$ library\footnote{\href{https://imgaug.readthedocs.io/en/latest/}{https://imgaug.readthedocs.io/en/latest/}}. The variations and resultant samples using both techniques can be found in Figure \ref{fig:samples}. \label{sec:vpd}

%% file: figures_tex/edge_extraction.tex
\begin{figure}[!ht]
\centering

\includegraphics[width=0.478275\textwidth]{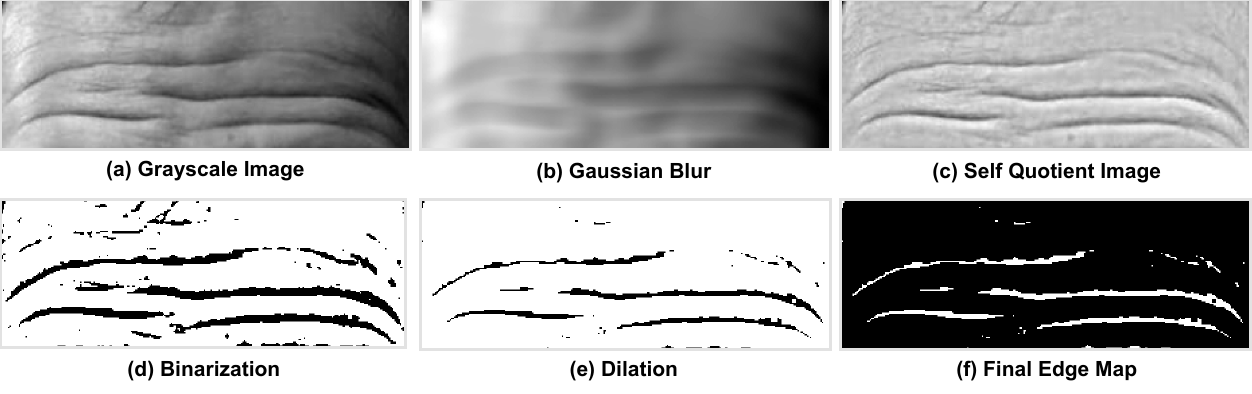}
\caption{Edge Extraction: A forehead-creases image at each pre-processing stage (a)-(f) to obtain the final edge map (f).}

\label{fig:edge_extraction}
\end{figure}

%% file: sections/5_experimental_setup.tex
\section{Experimental Setup} \label{label:experimental_setup}

In this section, we highlight the real datasets and various synthetic datasets sampled using our proposed method that are used in training and evaluating a forehead-creases verification system \cite{bharadwaj2022mobile} while following a cross-database verification protocol. We discuss our performance evaluation protocol, where we carry out three experiments to validate the robustness of our proposed method. 

\subsection{Real Datasets:} We use the real FH-V1 dataset \cite{bharadwaj2022mobile} with 247 subjects in our experiments (Table \ref{table:eer}), and to obtain training image pairs for the $Edge2FC$ model.  For evaluation, an additional dataset (FH-V2)$^{'}$ was introduced in \cite{tandon2024synthetic} that is used for cross-database verification with $29,200$ genuine and $8,497,200$ imposter matching pairs. Additional details about these datasets can be found in \cite{bharadwaj2022mobile} and \cite{tandon2024synthetic}.

\input{figures_tex/comparison}

\subsection{Synthetic Datasets:} We sample $247$ synthetic identities (for a fair comparison with the subject-agnostic SA-PermuteAug synthetic dataset proposed in \cite{tandon2024synthetic}). $10$ synthetic forehead-creases (FC) images are sampled using the $Edge2FC$ model for one input visual prompt with low variations. When visual prompts are perturbed to achieve control-point diversity (CPD), one image is sampled for each of the $10$ perturbed edge maps. Similarly, one image corresponding to each of the 14 augmentations (Section \ref{sec:vpd}) is sampled to achieve visual prompt diversity (VPD). The resulting synthetic databases used in all our evaluations are summarised in Table \ref{table:synthetic_datasets} with appropriate suffixes. For a fair comparison, the subject-specific synthetic databases in \cite{tandon2024synthetic} are not included, as they rely on real IDs to generate the corresponding synthetic poses and are not novel synthetic IDs.

\input{tables/2_synthetic_datasets}

Furthermore, we also train an unconditional DDPM \cite{dhariwal2021diffusion} on dilated self-quotient images to generate new edge maps that are translated using the $Edge2FC$ model (Prefix \textit{"DiffEdges"} are used in Tables \ref{table:eer} and \ref{table:fid} to denote these datasets). We also apply VPD to these edge maps, however, CPD cannot be applied on the same because these edge maps are static and lack local control, unlike b-splines.

\subsection{Forehead-Creases Verification System (FHCVS):}
We use FHCVS model \cite{bharadwaj2022mobile} as the backbone network, i.e. ResNet-18 \cite{he2016deep} followed by spatial \cite{fu2019dual} and channel attention modules \cite{wang2020eca}, and trained using AdaFace \cite{kim2022adaface} and focal loss \cite{lin2017focal}. 

\textbf{Training Curriculum:} We observe a significant gap between real and synthetic data (Table \ref{table:fid}) that affects the performance of FHCVS when both data are merged. To mitigate this gap and prevent the model from overfitting on synthetic data, we propose a \textit{training curriculum} that gradually increases the amount of synthetic data the recognition model is trained on. We start by fine-tuning the real baseline model and introduce a random subset of $50$ synthetic IDs along with real data. On convergence, training is continued for a new set of $50$ synthetic IDs, along with real and synthetic data of the previous cycle, and so on (Here $50$ is chosen to limit the total number of cycles and reduce computational complexity). Finally, the best-performing model of all cycles (in terms of EER), is fine-tuned on real data. \label{training_curriculum}

\subsection{Performance Evaluation Protocol}
We evaluate the performance of FHCVS \cite{bharadwaj2022mobile} to showcase the efficacy of our proposed synthetic databases by conducting the following three experiments: 
\begin{itemize}
    \item \textbf{Experiment 1:} This experiment leverages the Real (FH-V1) dataset and the Subject-Agnostic dataset \cite{tandon2024synthetic} to train the FHCVS model both with and without the proposed training curriculum. The results demonstrate the effectiveness of the curriculum in enhancing model performance. Once trained, the FHCVS model serves as a feature extractor to generate embeddings for probe and gallery images in the (FH-V2)$^{'}$ dataset, adhering to a cross-database matching protocol.
    \item \textbf{Experiment 2:}
    The proposed B-spline databases—comprising low-diversity (FC), control point diversity (CPD), and visual prompt diversity (VPD)—are each merged with the FH-V1 (real) dataset and used to train the FHCVS model while adhering to the proposed training curriculum. As demonstrated in Experiment 1, the absence of the training curriculum results in poor performance. Therefore, for consistency and simplicity, all subsequent experiments implement this strategy. This experiment underscores the effectiveness of incorporating visual prompt diversity in training. Similar to Experiment 1, the FHCVS model trained on synthetic data is utilized as a feature extractor for cross-database evaluation on the real (FH-V2)$^{'}$ dataset, further validating its robustness.
    \item \textbf{Experiment 3:} This experiment evaluates the effectiveness of our proposed method for sampling synthetic forehead creases by geometrically modeling them using B-splines, compared to directly estimating the underlying data distribution of forehead-crease edge maps. Interestingly, both methods demonstrate comparable performance. To further enhance results, we propose merging the two databases, which yields improved performance when evaluating the FHCVS on the (FH-V2)$^{'}$ dataset. These findings emphasize the robustness and effectiveness of the overall approach. 
\end{itemize}

%% file: figures_tex/comparison.tex
\begin{figure*}[!ht]
\centering

\includegraphics[width=0.95\textwidth]{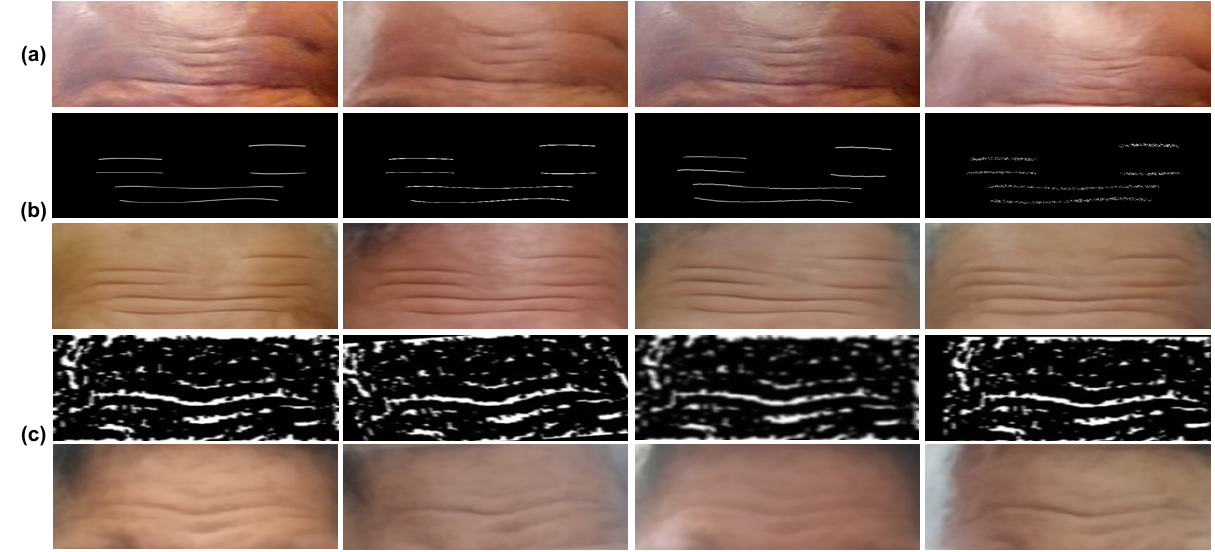}
\caption{Comparison: Synthetic IDs from (a) SA-PermuteAug \cite{tandon2024synthetic}, (b) BSpline-VPD, and,  (c) DiffEdges-VPD datasets. DiffEdges-VPD offers higher realism (Table \ref{table:fid}), however both offer comparable quantitative performance (Table \ref{table:eer}).}

\label{fig:comparison}
\end{figure*}

%% file: tables/2_synthetic_datasets.tex
\begin{table}[htbp]
    \centering  
     
    \begin{tabular}{ccccccc}
    \hline
    \textbf{Synthetic Dataset} & \textbf{Venue} & \textbf{Subjects} & \textbf{Images} \\ \hline
    
 SA-PermuteAug & IJCB'24 & 247 & 2717  \\ \hline
 BSpline-FC & Ours & 247 & 2470 \\
 BSpline-CPD & Ours & 247 & 2470 \\
 BSpline-VPD & Ours & 247 & 3458 \\ 
 \hline
 DiffEdges-FC & Ours & 247 & 2470 \\
 DiffEdges-VPD & Ours & 247 & 2470 \\
 \hline
    \end{tabular}
    \caption{Synthetic Databases used in our evaluations.}
   \label{table:synthetic_datasets}
\end{table}

%% file: sections/6_results.tex
\section{Experimental Results} \label{label:results}
In this section, we mention the evaluation metrics used in our experiments, present the qualitative performance of the synthetic datasets, the quantitative performance of FHCVS trained on real and synthetic datasets, and discuss our key observations.
\subsection{Evaluation Metrics}
To evaluate the quality of the synthetic forehead-creases images, we report the Fréchet Inception Distance (FID) score \cite{heusel2017gans}, intra-subject diversity \cite{batzolis2021conditional}, and Structural Similarity Index Measure (SSIM) \cite{wang2004image} in Table \ref{table:fid}. For quantitative evaluation of the FHCVS, we present the Equal Error Rate (EER), True-Match Rate (TMR) $@$ False-Match Rate (FMR) at $0.1\%$ and $0.01\%$ in Table \ref{table:eer}.

\input{tables/1_eer}

\subsection{Qualitative Evaluation}
The qualitative results are reported in Table \ref{table:fid} on each database mentioned in Table \ref{table:synthetic_datasets}. The FID score \cite{heusel2017gans} measures how close the synthetic forehead-creases images are to the real data distribution. To measure structural similarity, we take the first synthetic pose of an ID (that is generated without any variation), and measure its SSIM score with the remaining poses of the same ID. Finally, the average SSIM score is calculated over all the identities in the database.

Table \ref{table:fid} indicates an improved intra-subject diversity with visual prompt augmentations than control-point variations and outperforms $SA$ synthetic data proposed in \cite{tandon2024synthetic}, with a maximum of $19.198$ and $28.238$ for BSpline-VPD and DiffEdges-VPD databases respectively. This showcases the effectiveness of using VPD while CPD contributes less towards enhancing diversity within mated samples, which is also evident from Figure \ref{fig:samples}.
%%%%%%% FIGURE %%%%%%%%%

\input{tables/0_fid}

However, the results suggest a lack of realism in the synthetic samples generated using b-spines that obtain higher FID scores. The reason can be attributed to a lack of textures in visual prompts used during inference, while the $Edge2FC$ model is trained on texture-rich dilated self-quotient images. Improving textures to improve fidelity is essential \cite{jin2024pce} and can be addressed in future work.

\input{figures_tex/feature_maps}

DiffEdges databases (Figure \ref{fig:comparison}), although, do offer some competitive realism compared to the $SA$ dataset. In terms of structural similarity, we observe the expected trade-off between SSIM and diversity. Low diversity datasets (SA and FC) show higher SSIM than higher diversity datasets (VPD and CPD).

\subsection{Cross-Database Matching Results} \label{cross_db_results}

For each evaluation in Table \ref{table:eer}, a synthetic database is merged with the real FH-V1 dataset, thereby resulting in $247 \times 2 = 494$ identities that are utilized to train the FHCVS \cite{bharadwaj2022mobile}. First, the SA-PermuteAug dataset \cite{tandon2024synthetic} is used and the proposed training curriculum (Section \ref{training_curriculum}) is followed. We observe a drop in EER from $13.76\%$ to $10.53\%$ and an improvement in TMR@FMR=(0.1 / 0.01\%) rates from $33.62\%$ / $17.04\%$ to $51.43\%$ / $32.97\%$ when the training curriculum is used, outperforming the real baseline (EER $12.35\%$, TMR $40.12\%$ / $22.46\%$). This suggests the effectiveness of the proposed curriculum in bridging the gap between real and synthetic data while training. Since the method shows performance gains, the rest of the experiments follow the same strategy.

\input{figures_tex/det_curve}

Despite the lack of realism shown in Table \ref{table:fid}, BSpline databases outperform SA dataset, which was trained both with or without training curriculum, amongst which BSpline-VPD performs the best in terms of EER ($9.18\%$) and a TMR@FMR=0.1\% of $57.44\%$, while BSpline-FC performs the best with $41.83\%$ in terms of TMR@FMR=0.01\%. This can be attributed to the importance of creases in intermediate feature maps as shown in \cite{zhao2022bezierpalm} via squeeze-and-excite operations, also conclusive from intermediate feature maps Figure \ref{fig:feature_maps}. Moreover, this also highlights the importance of increased intra-subject diversity with visual prompt augmentations (Table \ref{table:fid}). 

\input{figures_tex/samples}

DiffEdges-VPD outperforms BSpline-VPD with a TMR of $58.49\%$ (at FMR=$0.1\%$) and $42.19\%$ (at FMR=$0.01\%$), while the later performing the best in terms of EER ($9.18\%$). Furthermore, we merge $50\%$ IDs from BSpline-VPD and DiffEdges-VPD databases respectively without shuffling, and the resultant database ($247$ synthetic IDs) along with the real FH-V1 data, outperforms every other training dataset with a TMR of $59.23\%$ (at FMR=$0.1\%$) and $43.63\%$ (at FMR=$0.01\%$). The training curriculum relies on gradual learning shifts from one dataset type to another, thereby enforcing the training from a less complex synthetic b-spline creases database towards synthetic images of higher fidelity sampled using unconditional diffusion. While we acknowledge geometrically modeling complex real-world forehead-creases images via parametric curves to generate realistic forehead-creases can be a challenging task, the results show that the FHCVS benefits from both, synthetic IDs sampled using b-splines that provide localized spatial control of creases, as well as creases learnt via generative modeling. The combined DET plot for all experiments is shown in Fig. \ref{fig:det}, while the t-SNE visualization comparing image embeddings of $1^{st}$ ten FH-V1 real IDs and BSpline-VPD synthetic IDs are presented in Figure \ref{tsne}.

%% file: tables/1_eer.tex
\begin{table*}[htbp]
    \centering
    \label{tab:my-table3}
    \begin{tabular}{cccccc}
    \hline
    \multirow{2}{*}{\textbf{Experiments}} & \multirow{2}{*}{\textbf{Train Database}} & 
    \multirow{2}{*}{\textbf{Train Curriculum}}  & \multirow{2}{*}{\textbf{EER (\%) \(\downarrow\)}} & \multicolumn{2}{c}{\textbf{TMR (\%) @ FMR (\%) =}} \\ \cline{5-6} 
     &  &  & & \textbf{0.1 (\%) \(\uparrow\)} & \textbf{0.01 (\%) \(\uparrow\)} \\ \hline

    & FH-V1 (Real)  & \ding{55} & 12.35 & 40.12 & 22.46  \\
    % & FH-V1 (Real)  & \ding{51} & 10.26 &  50.46 &  31.68 \\

    % ============ IJCB =============
    % FH-V1 (Real)         & 12.39  & 40.19 &  21.97 \\
    % FH-V1 with Translation   & 10.23  & 55.26 & 40.41 \\
    % FH-V1 with all Aug &  11.25 & 55.20 &  40.71   \\ \hline
    % SS-Comb & 11.83  & 49.94 & 34.42 \\
    % SS-Permute  &  10.37  & 56.06 & 40.57  \\
    % & FH-V1 + SS-Permute & 12.54 & 44.76 & 27.51 \\ 
    % SS-PermuteAug \cite{tandon2024synthetic} & 9.38 &  60.32 & 45.68  \\ \hline
    \multirow{1}{*}{\textbf{Experiment 1}} & Real + SA-PermuteAug \cite{tandon2024synthetic}  & \ding{55} & 13.76 & 33.62 & 17.04 \\ 
    
    % & FH-V1 + SS-Permute-Aug   & 13.26 & 45.23 & 29.39 \\
    % FH-V1 with all Aug + SS-Permute   & 10.46 & 57.69 & 41.95  \\ 
    % FH-V1 with all Aug + SS-PermuteAug & 11.20 &  56.13   & 41.78 \\  \hline
    
    & Real + SA-PermuteAug \cite{tandon2024synthetic} & \ding{51} & 10.53 & 51.43 & 32.97 \\ \hline
    \multirow{3}{*}{\textbf{Experiment 2}}& Real + BSpline-FC & \ding{51} & 10.03 & 57.06 & 41.83  \\
    & Real + BSpline-CPD & \ding{51} & 10.05 & 55.13 & 39.21 \\ 
    & Real + BSpline-VPD & \ding{51} & \textbf{9.18} & 57.44 & 40.39 \\ \hline
    % Real + BSpline-C2VD &  &  &  \\ \hline
    \multirow{3}{*}{\textbf{Experiment 3}} & Real + DiffEdges & \ding{51} & 9.49 & 56.73 & 38.87 \\
    & Real + DiffEdges-VPD & \ding{51} & 9.42 & 58.49 & 42.19 \\
    % & Real + (DiffEdges + BSpline)-VPD & \ding{51} & 9.65  & 56.87 & 38.90 \\\hline
    & Real + (DiffEdges + BSpline)-VPD & \ding{51} & 9.63  & \textbf{59.23} & \textbf{43.63} \\\hline

    \end{tabular}
    \caption{Quantitative results on Real (FH-V1 \cite{bharadwaj2022mobile}), subject-agnostic synthetic data (SA-PermuteAug \cite{tandon2024synthetic}) and synthetic IDs sampled using our proposed methods (Table \ref{table:synthetic_datasets}), when used as training databases. We use FHCVS model \cite{bharadwaj2022mobile} as the backbone with AdaFace \cite{kim2022adaface} and Focal Loss \cite{lin2017focal}. All trained models are used as feature extractors for cross-database evaluation on (FH-V2)$^{'}$ dataset.}
    \label{table:eer}
\end{table*}

% \begin{table*}[htbp]
%     \centering
    
%     \begin{tabular}{lccccc}
%     \hline
%     \multirow{2}{*}{\textbf{Experiments}} & \multirow{2}{*}{\textbf{Database}} & \multirow{2}{*}{\textbf{EER (\%) \(\downarrow\)}} & \multicolumn{2}{c}{\textbf{TMR (\%) @ FMR (\%) =}} \\ \cline{4-5} 
%      &  &  & \textbf{0.1 (\%) \(\uparrow\)} & \textbf{0.01 (\%) \(\uparrow\)} \\ \hline
     
%     \multirow{3}{*}{\textbf{Experiment 1}} & FH-V1         & 12.39  & 40.19 &  21.97 \\
%     & FH-V1 with Translation   & 10.23  & 55.26 & 40.41 \\
%     & FH-V1 with all Aug &  11.25 & 55.20 &  40.71   \\ \hline
%     \multirow{5}{*}{\textbf{Experiment 2}}& SS-Comb & 11.83  & 49.94 & 34.42 \\
%     & SS-Permute &  10.37  & 56.06 & 40.57  \\
%     & \textbf{SS-PermuteAug} & \textbf{9.38} &  \textbf{60.32} & \textbf{45.68} \\
%     & FH-V1 with all Aug + SS-Permute   & 10.46 & 57.69 & 41.95  \\ 
%     & FH-V1 with all Aug + SS-PermuteAug & 11.20 &  56.13   & 41.78 \\    \hline
%     \multirow{5}{*}{\textbf{Experiment 3}} & SS-BG-Permute &  14.92  & 25.05 & 10.33  \\
%     & SS-BG-PermuteAug & 17.67 & 22.93 & 9.68 \\
%     & FHV1 with all Aug + SS-BG-Permute & 11.06 &  57.09 & 43.81 \\
%     & FHV1 with all Aug + SS-BG-PermuteAug &  11.78 & 53.97  & 39.62  \\
    
%     \hline
    
%     \end{tabular}
%     \caption{Quantitative results on FH-V1 and Subject-Specific (SS) synthetic datasets.}
%     \label{table:eer}
% \end{table*}

%% file: tables/0_fid.tex
\begin{table}[htbp]
    \centering
    \begin{tabular}{ccccccc}
    \hline
    \textbf{Synthetic Dataset} & \textbf{FID}  \(\downarrow\) & \textbf{Diversity} \(\uparrow\) & \textbf{SSIM} \(\uparrow\) \\ \hline
    SA-PermuteAug \cite{tandon2024synthetic} & \textbf{55.999} & 11.236 & 0.834 \\ \hline
    BSpline-FC & 184.334 & 15.679 & 0.829 \\
    BSpline-CPD & 185.001 & 15.962 &  0.663 \\ 
    BSpline-VPD & 162.709 & 19.198 &  0.739  \\ \hline
    DiffEdges-FC & 76.984 & 13.558 &   \textbf{0.926} \\
    DiffEdges-VPD & 67.776 & \textbf{28.238} & 0.796  \\
    \hline
    \end{tabular}
    \caption{Qualitative Evaluation of Synthetic Datasets}
    \label{table:fid}
\end{table}

%% file: figures_tex/feature_maps.tex
\begin{figure}[!ht]
\centering

\includegraphics[width=0.478275\textwidth]{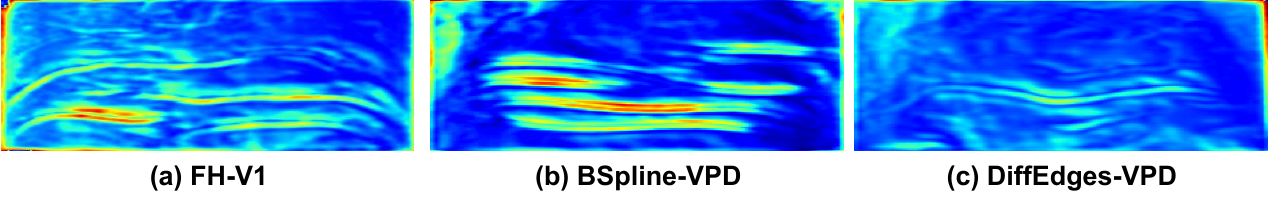}
\caption{Intermediate feature maps visualization for images from (a) FH-V1 (Real), (b) BSpline-VPD, and, (c) DiffEdges-VPD datasets. Despite high FID scores, the verification model mainly focuses on creases rendered using b-splines in (b), while both creases and textures in (a) and (c). Features maps resized from $(112, 112)$ for a better visualization.}

\label{fig:feature_maps}
\end{figure}

%% file: figures_tex/det_curve.tex
\begin{figure}[htbp]
\centering
\includegraphics[width=0.45\textwidth]{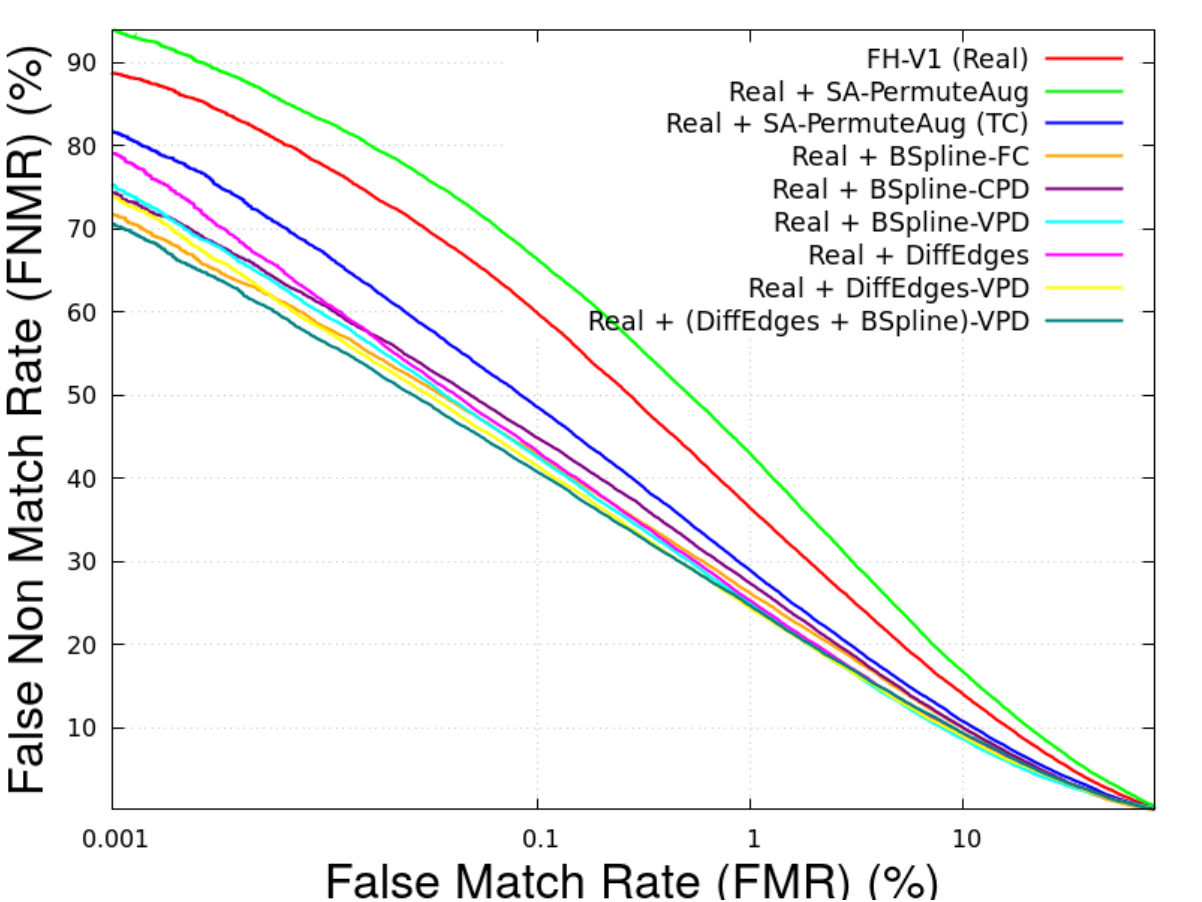}

\caption{Combined DET curve for comparative analysis among all different experiments.The x-axis is in log scale.}
\label{fig:det}
\end{figure}

%% file: figures_tex/samples.tex
\begin{figure}[!ht]
\centering

\includegraphics[width=0.475\textwidth]{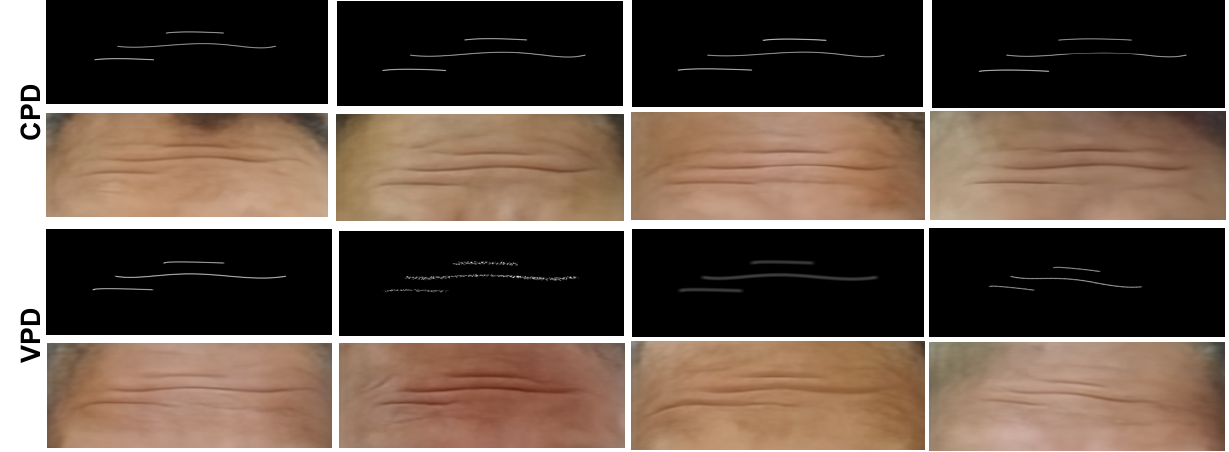}
\caption{Mated samples from a synthetic ID showcasing control point diversity (CPD) and visual prompt diversity (VPD). VPD offers higher intra-subject diversity than CPD (Table \ref{table:fid}), enhancing the verification performance of FHCVS (Table \ref{table:eer}).}

\label{fig:samples}
\end{figure}

%% file: sections/7_conclusion.tex
\section{Limitations} \label{label:limitations}
As discussed in Section \ref{cross_db_results}, geometrically rendering forehead creases does not fully capture the complexity of real-world images and requires substantial domain knowledge, especially given that forehead creases are a relatively new biometric modality. Additionally, the proposed approach does not effectively handle textures and skin tones, reducing the overall fidelity of the synthetic samples in some extreme cases. Future research could focus on refining texture patterns in the B-spline-based visual prompts and conditioning the $Edge2FC$ model to produce consistent skin colors for a given synthetic identity. This may lead to more realistic synthetic samples and improved model performance. 

\input{figures_tex/tsne}

\section{Conclusion} \label{label:conclusion}

We introduce a generalized parametric forehead-crease model that employs B-spline and Bézier curves on a dynamically generated grid mask, operating under well-defined constraints to sample novel edge maps (visual prompts). These prompts are then used to produce new synthetic IDs through a diffusion-based Edge-to-ForeheadCreases (Edge2FC) image translation generative network. In addition, we train an unconditional DDPM on dilated self-quotient images to generate further edge maps for the Edge2FC model. To enhance intra-subject diversity, we apply visual prompt augmentations to the edge maps before synthesizing new IDs. Our experimental results indicate that a forehead-crease verification model trained on both synthetic and real data achieves improved equal error rates and true matching rates, surpassing the baseline model trained solely on real identities. This demonstrates the efficacy of our approach in enriching the training data and ultimately strengthening the verification process.

%% file: figures_tex/tsne.tex
\begin{figure}[!ht]
\centering
\includegraphics[width=0.9\columnwidth]{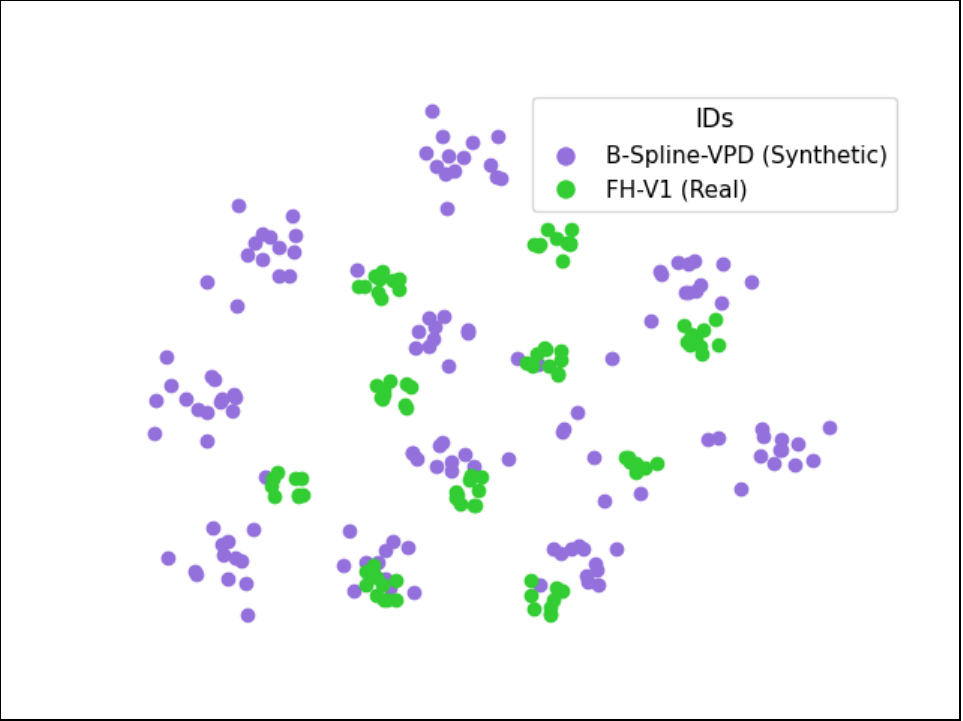}
\caption{t-SNE visualization: Real (FH-V1) and Synthetic (BSpline-VPD) image embeddings for $1^{st}$ ten IDs.}
\label{tsne}
\end{figure}

%% file: main.bbl
\begin{thebibliography}{10}\itemsep=-1pt

\bibitem{alhallak2024optimizing}
Kamal Alhallak.
\newblock Optimizing botulinum toxin a administration for forehead wrinkles: Introducing the lines and dots (lads) technique and a predictive dosage model.
\newblock {\em Toxins}, 16(2):109, 2024.

\bibitem{anido2017tailored}
Javier Anido, Daniel Arenas, Cristina Arruabarrena, Alfonso Dom{\'\i}nguez-Gil, Carlos Fajardo, Mar Mira, Javier Murillo, Natalia Rib{\'e}, Helga Rivera, Sofia Ruiz~del Cueto, et~al.
\newblock Tailored botulinum toxin type a injections in aesthetic medicine: consensus panel recommendations for treating the forehead based on individual facial anatomy and muscle tone.
\newblock {\em Clinical, cosmetic and investigational dermatology}, pages 413--421, 2017.

\bibitem{batzolis2021conditional}
Georgios Batzolis, Jan Stanczuk, Carola-Bibiane Sch{\"o}nlieb, and Christian Etmann.
\newblock Conditional image generation with score-based diffusion models.
\newblock {\em arXiv preprint arXiv:2111.13606}, 2021.

\bibitem{bharadwaj2022mobile}
Rohit Bharadwaj, Gaurav Jaswal, Aditya Nigam, and Kamlesh Tiwari.
\newblock Mobile based human identification using forehead creases: Application and assessment under covid-19 masked face scenarios.
\newblock In {\em Proceedings of the IEEE/CVF Winter Conference on Applications of Computer Vision}, pages 3693--3701, 2022.

\bibitem{deng2009imagenet}
Jia Deng, Wei Dong, Richard Socher, Li-Jia Li, Kai Li, and Li Fei-Fei.
\newblock Imagenet: A large-scale hierarchical image database.
\newblock In {\em 2009 IEEE conference on computer vision and pattern recognition}, pages 248--255. Ieee, 2009.

\bibitem{dhariwal2021diffusion}
Prafulla Dhariwal and Alexander Nichol.
\newblock Diffusion models beat gans on image synthesis.
\newblock {\em Advances in neural information processing systems}, 34:8780--8794, 2021.

\bibitem{fu2019dual}
Jun Fu, Jing Liu, Haijie Tian, Yong Li, Yongjun Bao, Zhiwei Fang, and Hanqing Lu.
\newblock Dual attention network for scene segmentation.
\newblock In {\em Proceedings of the IEEE/CVF conference on computer vision and pattern recognition}, pages 3146--3154, 2019.

\bibitem{goodfellow2020generative}
Ian Goodfellow, Jean Pouget-Abadie, Mehdi Mirza, Bing Xu, David Warde-Farley, Sherjil Ozair, Aaron Courville, and Yoshua Bengio.
\newblock Generative adversarial networks.
\newblock {\em Communications of the ACM}, 63(11):139--144, 2020.

\bibitem{10744493}
Steven~A. Grosz and Anil~K. Jain.
\newblock Genpalm: Contactless palmprint generation with diffusion models.
\newblock In {\em 2024 IEEE International Joint Conference on Biometrics (IJCB)}, pages 1--9, 2024.

\bibitem{he2016deep}
Kaiming He, Xiangyu Zhang, Shaoqing Ren, and Jian Sun.
\newblock Deep residual learning for image recognition.
\newblock In {\em Proceedings of the IEEE conference on computer vision and pattern recognition}, pages 770--778, 2016.

\bibitem{heusel2017gans}
Martin Heusel, Hubert Ramsauer, Thomas Unterthiner, Bernhard Nessler, and Sepp Hochreiter.
\newblock Gans trained by a two time-scale update rule converge to a local nash equilibrium.
\newblock {\em Advances in neural information processing systems}, 30, 2017.

\bibitem{jia2008palmprint}
Wei Jia, De-Shuang Huang, and David Zhang.
\newblock Palmprint verification based on robust line orientation code.
\newblock {\em Pattern Recognition}, 41(5):1504--1513, 2008.

\bibitem{jin2024pce}
Jianlong Jin, Lei Shen, Ruixin Zhang, Chenglong Zhao, Ge Jin, Jingyun Zhang, Shouhong Ding, Yang Zhao, and Wei Jia.
\newblock Pce-palm: Palm crease energy based two-stage realistic pseudo-palmprint generation.
\newblock In {\em Proceedings of the AAAI Conference on Artificial Intelligence}, volume~38, pages 2616--2624, 2024.

\bibitem{kim2022adaface}
Minchul Kim, Anil~K Jain, and Xiaoming Liu.
\newblock Adaface: Quality adaptive margin for face recognition.
\newblock In {\em Proceedings of the IEEE/CVF Conference on Computer Vision and Pattern Recognition}, 2022.

\bibitem{kumar2015recovering}
Ajay Kumar and Bichai Wang.
\newblock Recovering and matching minutiae patterns from finger knuckle images.
\newblock {\em Pattern Recognition Letters}, 68:361--367, 2015.

\bibitem{li2023bbdm}
Bo Li, Kaitao Xue, Bin Liu, and Yu-Kun Lai.
\newblock Bbdm: Image-to-image translation with brownian bridge diffusion models.
\newblock In {\em Proceedings of the IEEE/CVF conference on computer vision and pattern Recognition}, pages 1952--1961, 2023.

\bibitem{lin2017focal}
Tsung-Yi Lin, Priya Goyal, Ross Girshick, Kaiming He, and Piotr Doll{\'a}r.
\newblock Focal loss for dense object detection.
\newblock In {\em Proceedings of the IEEE international conference on computer vision}, pages 2980--2988, 2017.

\bibitem{moon2022fine}
Taehong Moon, Moonseok Choi, Gayoung Lee, Jung-Woo Ha, and Juho Lee.
\newblock Fine-tuning diffusion models with limited data.
\newblock In {\em NeurIPS 2022 Workshop on Score-Based Methods}, 2022.

\bibitem{patrikalakis2002shape}
Nicholas~M Patrikalakis and Takashi Maekawa.
\newblock {\em Shape interrogation for computer aided design and manufacturing}, volume~15.
\newblock Springer, 2002.

\bibitem{rombach2022high}
Robin Rombach, Andreas Blattmann, Dominik Lorenz, Patrick Esser, and Bj{\"o}rn Ommer.
\newblock High-resolution image synthesis with latent diffusion models.
\newblock In {\em Proceedings of the IEEE/CVF conference on computer vision and pattern recognition}, pages 10684--10695, 2022.

\bibitem{10593960}
Geetanjali Sharma, Gaurav Jaswal, Aditya Nigam, and Raghavendra Ramachandra.
\newblock Fh-sstnet: Forehead creases based user verification using spatio-spatial temporal network.
\newblock In {\em 2024 12th International Workshop on Biometrics and Forensics (IWBF)}, pages 1--6, 2024.

\bibitem{shen2023rpg}
Lei Shen, Jianlong Jin, Ruixin Zhang, Huaen Li, Kai Zhao, Yingyi Zhang, Jingyun Zhang, Shouhong Ding, Yang Zhao, and Wei Jia.
\newblock Rpg-palm: Realistic pseudo-data generation for palmprint recognition.
\newblock In {\em Proceedings of the IEEE/CVF International Conference on Computer Vision}, pages 19605--19616, 2023.

\bibitem{tandon2024synthetic}
Abhishek Tandon, Geetanjali Sharma, Gaurav Jaswal, Aditya Nigam, and Raghavendra Ramachandra.
\newblock Synthetic forehead-creases biometric generation for reliable user verification.
\newblock In {\em 2024 IEEE International Joint Conference on Biometrics (IJCB)}, pages 1--9. IEEE, 2024.

\bibitem{wang2020eca}
Qilong Wang, Banggu Wu, Pengfei Zhu, Peihua Li, Wangmeng Zuo, and Qinghua Hu.
\newblock Eca-net: Efficient channel attention for deep convolutional neural networks.
\newblock In {\em Proceedings of the IEEE/CVF conference on computer vision and pattern recognition}, pages 11534--11542, 2020.

\bibitem{wang2004image}
Zhou Wang, Alan~C Bovik, Hamid~R Sheikh, and Eero~P Simoncelli.
\newblock Image quality assessment: from error visibility to structural similarity.
\newblock {\em IEEE transactions on image processing}, 13(4):600--612, 2004.

\bibitem{zhao2022bezierpalm}
Kai Zhao, Lei Shen, Yingyi Zhang, Chuhan Zhou, Tao Wang, Ruixin Zhang, Shouhong Ding, Wei Jia, and Wei Shen.
\newblock B{\'e}zierpalm: A free lunch for palmprint recognition.
\newblock In {\em European Conference on Computer Vision}, pages 19--36. Springer, 2022.

\end{thebibliography}
